%% file: main.tex
\documentclass[letterpaper]{article} 
\usepackage{aaai23}  
\usepackage{times}  
\usepackage{helvet}  
\usepackage{courier}  
\usepackage[hyphens]{url}  
\usepackage{graphicx} 
\usepackage{natbib}  
\usepackage{caption} 
\usepackage{algorithm}
\usepackage{algorithmic}

\usepackage{newfloat}
\usepackage{listings}
\DeclareCaptionStyle{ruled}{labelfont=normalfont,labelsep=colon,strut=off} 
\floatstyle{ruled}
\newfloat{listing}{tb}{lst}{}
\floatname{listing}{Listing}
\nocopyright

\usepackage{caption}
\usepackage{subcaption}
\usepackage{svg}
\usepackage{multirow}
\title{Gender mobility in the labor market with skills-based matching models 
}

\author{
    Ajaya Adhikari\textsuperscript{\rm 1},
    Steven Vethman\textsuperscript{\rm 1},
    Daan Vos\textsuperscript{\rm 1},
    Marc Lenz\textsuperscript{\rm 1},
    Ioana Cocu\textsuperscript{\rm 1},
    Ioannis Tolios\textsuperscript{\rm 1},
    Cor J. Veenman\textsuperscript{\rm 1,2}
}
\affiliations{
    \textsuperscript{\rm 1}Netherlands Organisation for Applied Scientific Research, The Hague, The Netherlands\\
    \textsuperscript{\rm 2}Leiden University, Leiden, The Netherlands
}

\usepackage{bibentry}

\begin{document}

\maketitle
\input{chapters/abstract.tex}
\input{chapters/introduction.tex}
\input{chapters/related_work.tex}
\input{chapters/dataset.tex}

\input{chapters/experiments.tex}

\input{chapters/conclusion.tex}
\input{chapters/acknowledments.tex}

\bibliography{main}

\end{document}

%% file: chapters/abstract.tex
\begin{abstract}
Skills-based matching promises mobility of workers between different sectors and occupations in the labor market. 
In this case, job seekers can look for jobs they do not yet have experience in, but for which they do have relevant skills. 
Currently, there are multiple occupations with a skewed gender distribution.
For skills-based matching, it is unclear if and how a shift in the gender distribution, which we call \emph{gender mobility}, between occupations will be effected.
It is expected that the skills-based matching approach will likely be data-driven, including computational language models and supervised learning methods.

This work, first, shows the presence of gender segregation in language model-based skills representation of occupations.
Second, we assess the use of these representations in a potential application based on simulated data, and show that the gender segregation is propagated by various data-driven skills-based matching models.
These models are based on different language representations (bag of words, word2vec, and BERT), and distance metrics (static and machine learning-based).
Accordingly, we show how skills-based matching approaches can be evaluated and compared on matching \emph{performance} as well as on the risk of \emph{gender segregation}.
Making the gender segregation bias of models more explicit can help in generating healthy trust in the use of these models in practice.

\end{abstract}

%% file: chapters/introduction.tex
\section{Introduction}
The skills-based matching approach aims to create a better fit between those looking for a job (candidates) and those offering a job (employers) by means of focusing on the skills necessary to be successful at the job.
Previous research has shown that job satisfaction of employees increases, when one's skills are well matched with the job activities \cite{vieira2005skill}.
Next, matching based on skills  has the potential to allow more labor mobility of job seekers.
The focus on skills stimulates the formulation of which tasks the job seekers can perform, instead of diplomas obtained or professional experiences in similar functions in a specific sector. 
The emphasis on what job seekers can do, relative to what they have previously experienced, allows for more job opportunities (possibly in a new sector) for job seekers as well as a bigger pool of applicants for employers, especially in bottleneck occupations.
The potential of this approach has been recognized by various organizations through different initiatives \cite{prevoo2022inzicht, rentzsch2020skills}.

Gender segregation is largely present in the labor market \cite{bls, bettio2009gender, tomaskovic2006documenting}.
Previous research has shown that the increase in diversity in teams is associated with increased innovation \cite{nathan2013cultural, diaz2013gender}, effectiveness \cite{phillips2009pain}, productivity \cite{herring2009does}, and fair decision making \cite{sommers2006racial}. 
Moreover, the increase of diversity in occupations with a good representation of different groups, can serve social justice goals for those  groups that were historically banned or discouraged to practice certain occupations \cite{eagly2016passionate}.
Although labor mobility of workers may increase with skills-based matching, it is unclear whether a shift in gender distribution between occupations, which we call \emph{gender mobility}, will be effected.

Recruitment using skills-based matching is most probably to be data-driven \cite{prevoo2022inzicht} in which computational language models are used to interpret and represent skills formulation in machine readable vectors.
Using these vectors, machine learning has the opportunity to learn from historical candidate-job matches which skill sets of candidates fit well with the required skill sets of job openings.

This work investigates the effect of data-driven skills-based matching models on the gender segregation in occupations.
We first examine whether gender segregation exists in different skills representations (bag of words 
\cite{li2020survey}, word2vec \cite{word2vec}, and Bert \cite{reimers-2019-sentence-bert}) of occupations.
Second, we measure the impact of biases in skills representation on gender segregation for the application of skills-based matching based on simulated data. 
Typically, these biases are only studied in the language representations \cite{delobelle2022measuring,goldfarb-tarrant-etal-2021-intrinsic} without taking the effect on the possible down stream tasks into account.
We examine various skills-based matching models based on the above mentioned type of representations and different distance metrics (static and machine learning-based).
This allows the evaluation of a model not only according to its performance but also on its effect on the propagation of gender segregation.

This paper is structured as follows. Section 2 elaborates on the related work. In section 3, we describe the dataset and the preprocessing steps that were done. Section 4 describes the experiments and the outcomes. At last, we conclude in section 5.

%% file: chapters/related_work.tex
\section{Related work}\label{sec:background}

To put our contribution into perspective, we first highlight related work on gender segregation in the context of skills. 
Since we consider skills formulations as one of the sources of the segregation as well as the historical presence of genders in the respective jobs, we also review the literature on bias measurement with language models. 

We follow related work by defining gender segregation as under (over) representation of a given gender in occupations \cite{bettio2009gender}. 
One study showed that the female-dominated occupations are often associated with lower salary than male-dominated occupations, which partly stems from low valuation and visibility of female-associated skills \cite{grimshaw2007undervaluing}. 
Academics also point towards the fact that the type of skills matter.
In countries with more focus on general skills rather than role- or firm-specific skills less occupational gender segregation is present \cite{estevez2005gender}. 
Here, general skills are argued to be more gender-neutral as women have higher concern for skill portability given potential career interruptions due to family responsibilities. 
In the U.S., patterns of skills demand signal an increase in gender segregation that reinforces gender income inequality \cite{bettio2009gender}.
This is given that the shifts in demand towards lower paid work concern female-associated skills and shifts towards higher paid work concern male associated skills.
 In Britain, evidence was found that skills commonly acquired by women are increasingly in demand in highly-skilled occupations, such that gender segregation in those occupations may decrease \cite{brynin2016gender}. 
 All in all, the interrelation of skills and gender segregation have been established in related work, where the level and type of skills one has may impact gender segregation. 
 We contribute by investigating the interplay of skills and gender segregation within the scope of how risks for this bias can be measured within AI and language model applications that may foster skills-based matching.   

These language models are based on our written texts which reflect the stereotypes and human biases that are embedded in our language \cite{menegatti2017gender, caliskan2017semantics}. The field of trustworthy AI \cite{thuraisingham2022trustworthy} has therefore established a large knowledge base on measuring the bias in these models \cite{hovy2016social}. 
We acknowledge the recent criticism that relevant bias measurement requires the context of the downstream task in which the language model is applied, i.e. a shift from measuring language bias in the models themselves towards measuring the impact language bias has when language models are used in e.g. decision support tools \cite{delobelle2022measuring,goldfarb-tarrant-etal-2021-intrinsic}. 
In our contribution, we therefore measure the risk of language bias exacerbating gender segregation in the context of the application of skills-based matching. 

%% file: chapters/dataset.tex
\section{Dataset} \label{sec:dataset}

We position our work in the framework of skills of the O*NET database \cite{onet}. 
The O*NET database provides a structured hierarchy of sectors to occupations and their corresponding required skills.
The skills described as \emph{detailed work activities} are chosen for the experiments, as they have a similar level of tangibility in describing the required skills as is observed in job vacancies. 

To assess the gender segregation we also need the gender distribution of the O*NET occupations. 
The average gender distribution per O*NET occupation is retrieved from the U.S. Bureau of Labor Statistics (BLS) \cite{bls}. 

We would, ideally, need real historic matching data between the skills of job seekers and skills in job ads such that we can train and test skills-based matching models.
As skills-matching is not yet widespread and known initiatives protect their data for privacy reasons, this data is not available. 
For this reason, we gauge the potential impact of skills matching models on simulated matches of candidates and job openings based on skill profiles of O*net occupations.
Two random sub-sets of skills sampled from the same occupation are regarded as a good match while two random sub-set of skills sampled from two different occupations are regarded as a bad match. 
For the experiments, we choose to sample sub-sets of size 5. 
The ability of a candidate is present in the combination of the different skills as a whole.
One possible way to materialize this in the experiment is by concatenating the skill sets into one string.
This results in a dataset of good and bad matches between two skills descriptions.
3940 amount of pairs were created with equal amount of good and bad matches.
We choose for the (possible) occurrence of overlap in sampling of two sub-sets of skills from an occupation; this is also likely to be the case in reality as people/job openings will overlap in skill.
This set is further equally spit into a train and a test set.
For both the train and test we draw skill profiles from all occupations while the specific skills of different occupations do not overlap between the train and test set. 
This is also the realistic scenario, as the model in practice would also train and test on skill sets of candidates and job openings related to the same occupations.
To create a fair test set, some sub-sets of skills of a baker, for example, is trained upon, while the model is tested on other subsets of the the total skill set of a baker.

%% file: chapters/experiments.tex
\section{Experiments}\label{sec:experiments}
This section describes the setup and results of two experiments: (1) presence of gender segregation in occupations according to their skills representation, and (2) the evaluation of skills-based matching models according to their performance and the risk of propagating gender segregation.

For both experiments, we consider three types of vectorizers with different levels of complexity to convert a string of skills descriptions into one vector, namely Bag of Words (BoW) \cite{li2020survey}, word2vec \cite{word2vec} and Sentence-BERT \cite{reimers-2019-sentence-bert}.
BoW counts the presence of single or combination of words, while word2vec and Sentence-BERT are language models which encode the semantic meaning of a word in a vector.
Word2vec encodes each word of a sentence independently while Sentence-BERT also encodes the specific sentence context of a word.

All experiments were run in python 3.9.12. 
The BoW vectorization is implemented using 1-gram and 2-gram words in Scikit-learn 1.1.1. library \cite{scikit-learn}.
For word2vec, the vectors of different words are averaged to create one vector using the Spacy 3.4.1.
 \cite{spacy2} library. 
At last, the sentence transformers 1.2.1. library (implementation of Sentence-BERT) is used, which can transform a description of skills into one vector.

\subsection{Gender segregation in occupations based on skills}
\begin{figure*}
     \centering
     \begin{subfigure}[b]{0.33\textwidth}
         \centering
         \includegraphics[width=\textwidth]{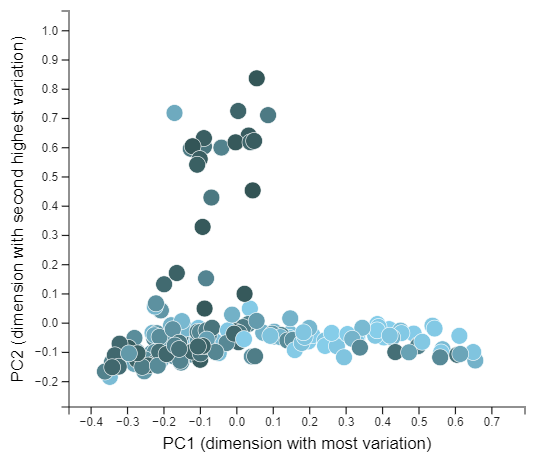}
         \caption{BoW}
         \label{Bag of Words}
     \end{subfigure}
     \hfill
     \begin{subfigure}[b]{0.33\textwidth}
         \centering
         \includegraphics[width=\textwidth]{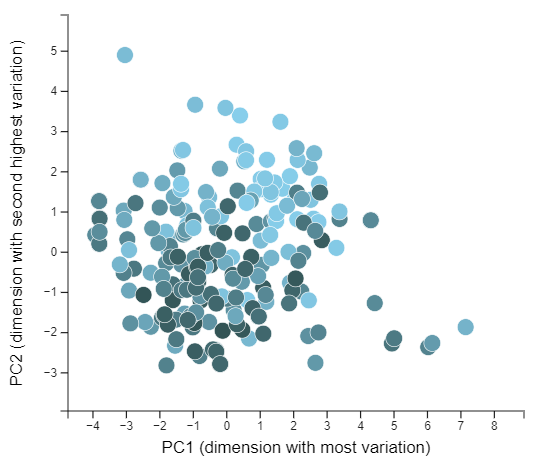}
         \caption{Word2vec}
         \label{Word2vec}
     \end{subfigure}
     \hfill
     \begin{subfigure}[b]{0.33\textwidth}
         \centering
         \includegraphics[width=\textwidth]{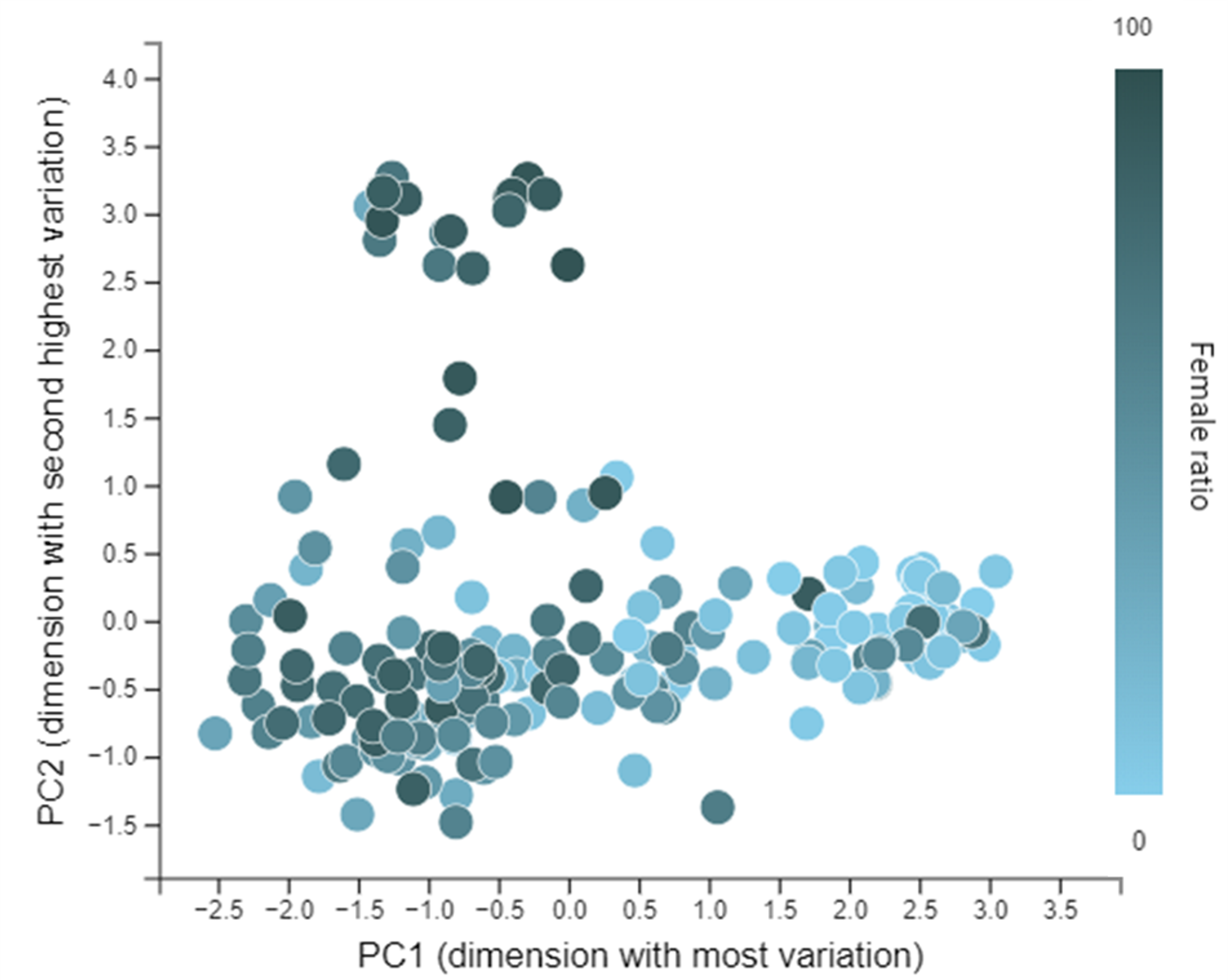}
         \caption{Sentence-BERT}
         \label{BERT}
     \end{subfigure}
        \caption{These figures visualize the current gender segregation in occupations based on their skills (represented by 3 types of vectorizers)\textsuperscript{\ref{footnote:webversion}}.}
        \label{fig:pca_visualization}
\end{figure*}

We investigate the current gender segregation in occupations according to their underlying skills.
The list of skills of each O*NET occupation is concatenated together to create one large string per occupation.
These strings are fed to the three above mentioned vectorizers.
The resulting 3 vectors per occupation are further separately visualized in Figure \ref{fig:pca_visualization}\footnote{\label{footnote:webversion}The interactive web version of the results of both experiments can be found in \url{ https://fate2022-demo.tnodatalab.nl/#/model}} by mapping them into two dimensions using PCA, which encodes the variance in the data as much as possible.
Each point represents an occupation, and the color indicates the average female gender ratio of that occupation according to BLS.

For all three types of vectorizers, we see clusters of occupations according to their gender ratio.
For example, Sentence-BERT shows a cluster of male majority occupations on the middle right and a female majority occupations on top left.
The male majority cluster contains mostly technical or labor intensive occupations such as machinist, dishwasher and roofers, while the female majority cluster contains mostly healthcare occupations such as dental assistants, registered nurse and respiratory therapists.
Thus, this experiment indicates that there is a risk that language models interpreting the skills sets of occupations are influenced by gender segregation.  

\begin{table}[]
\centering
\begin{tabular}{llll}
\hline
Vectorizer                    & Similarity/Distance Metr. & AUC & GSR \\ \hline
\multirow{3}{*}{BoW} & Metric learning             & 0,90             & 0,60                     \\
                              & Cosine                      & 0,94             & 0,72                   \\
                              & Euclidean                   & 0,88             & 0,59                   \\ \hline
\multirow{3}{*}{Word2Vec}     & Metric learning             & 0,91             & 0,68                    \\
                              & Cosine                      & 0,82              & 0,57                   \\
                              & Euclidean                   & 0,83             & 0,60                     \\ \hline
\multirow{3}{*}{Sentence-BERT}         & Metric learning             & 0,91             & 0,69                    \\
                              & Cosine                      & 0,94             & 0,70                   \\
                              & Euclidean                   & 0,90               & 0,67                   
\end{tabular}
\caption{This table shows the trade-off between AUC performance and GSR for three types of vectorizers (BoW, Word2vec and Sentence-BERT embedding) in combination with three types of similarity/distance metrics (Euclidean distance, Cosine similarity and Metric learning)\textsuperscript{\ref{footnote:webversion}}.}
\label{table:gender_segregation_propagation}
\end{table}

\subsection{Evaluation of skills-based matching models}
In the second experiment, we put the perceived risk for gender segregation in the language representation of skills from the first experiment into the relevant context. 
In particular, we demonstrate how skills-matching models can be evaluated not only for their matching performance but also for their risk of propagating gender segregation.
We demonstrate this setting with three types of the above mentioned vectorizers and three types of similarity measures namely Euclidean distance, cosine similarity and metric learning. 
This results in 9 versions of skills-based matching models.
While Euclidean and Cosine metrics are pre-defined, metric learning is a supervised learning method which learns a task-specific similarity measure from the training data. 
The \emph{Information Theoretic Metric Learning} model with default parameters of the metric-learn 0.6.2 library\cite{metric-learn} is used in this experiment.

The dataset containing skills description pairs of good and bad matches described in section \ref{sec:dataset} is used to train and test the skills-based matching models.
Each pair of skills descriptions of the test set is first vectorized and the resulting two vectors are further provided as input to a distance or similarity metric resulting in a matching score.
The performance evaluation of the 9 versions according to the AUC metric can be found in Table \ref{table:gender_segregation_propagation}\textsuperscript{\ref{footnote:webversion}}.
We didn't further optimize the metric learning algorithm as it is not the focus of this work.

\begin{figure}[t]
\includegraphics[scale=0.58]{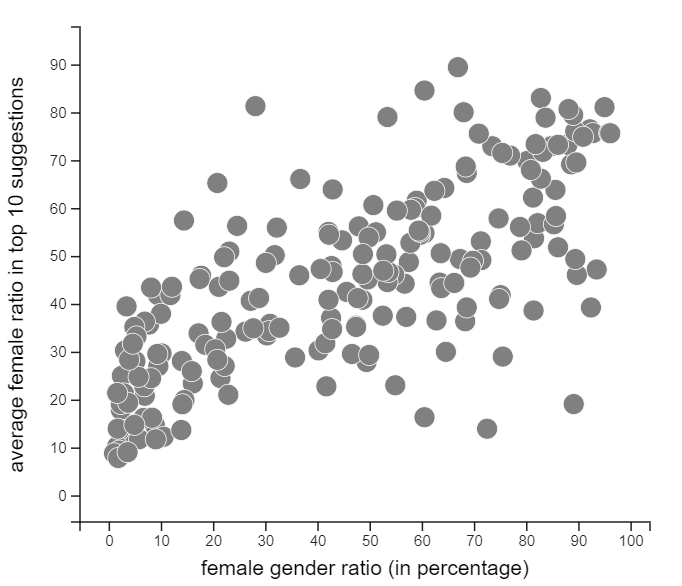}
\centering
\caption{The figure illustrates the female gender ratio of an occupation and the average female ratio of 10 occupations with the highest matching score according to the BoW and cosine similarity version.}
\label{fig:example_gsr}
\end{figure}

To assess the risk of propagating gender segregation by the above mentioned skills-based matching models, we create a new test-set.
Similar to the training data, a random sub-set of 5 skills of each O*NET occupation is extracted and concatenated together.
Per O*NET occupation the matching score is computed with the rest of the occupations.
For each occupation, we find the 10 occupations with the highest matching score with respect to their skill set. 
Then, the overall \emph{Gender Segregation Risk (GSR)} of a skills-based matching model is measured by the Pearson correlation  between the following variables: the female gender ratio of an occupation and the average female ratio of 10 occupations with the highest matching score.
An example of the correlation between these two variables for the BoW and cosine similarity version is shown in Figure \ref{fig:example_gsr}.
This score provides an indication of how likely job seekers will end up in a new occupation with similar gender distribution as their previous occupation if they follow one of the top 10 suggestions made by the matching model.
Table \ref{table:gender_segregation_propagation}\textsuperscript{\ref{footnote:webversion}} shows the GSR of the different matching models.
Note that in this experiment, it is not an issue that there might be overlap between the train-set and this new test-set, because the model is not optimized to minimize the GSR.

We see a positive correlation of higher than 0.5 for all models, indicating that gender segregation is propagated by these models.
We also see that this risk is correlated with the performance of the model.
This is expected because, for example, a bad model which randomly assigns a matching score, will randomly suggest the top 10 occupations.
The gender ratio of this suggested occupations will not be correlated with the query occupation.

This allows one to evaluate and compare different models according to its performance and the GSR.
In our simulated application, the combination of BoW and cosine has the best performance.
While the BoW with metric learning performs slightly less, it has the best trade-off between high performance and low GSR.

%% file: chapters/conclusion.tex
\section{Conclusion}\label{sec:conclusion}
We measure the risk for gender segregation in skills descriptions and next to that demonstrate with simulated data the risk of propagating gender segregation within a potential skills-matching application that uses language models to interpret skills descriptions. 
The measured language bias in skills has shown the need to consider the risk for gender segregation when language models are put into use for skills-based matching.
To facilitate this, our work provides a first exploration on how quantifying the measured risk for gender segregation can aid design choices in practice by considering both the performance and the risk of propagating gender segregation when choosing a skills-based matching model.
The quantification of this bias can contribute towards generating healthy trust in the use of skills-based matching models.

Future work is needed with real data to validate to what extent this risk is also present in applications and to establish how this risk can be integrated in design choices.
Moreover, future research is advised to include more diversity aspects such as ethnicity and educational background.

%% file: chapters/acknowledments.tex
\section*{Acknowledgements}
This work was part of FATE and Skills-matching projects which were funded by the Appl.AI program within TNO.